\documentclass[runningheads,a4paper]{llncs}
\usepackage{hyperref}
\usepackage{amssymb}
\usepackage{graphicx}
\usepackage{cleveref}
\usepackage{amsmath} 
\usepackage{booktabs}
\usepackage{tabularx}
\usepackage{subcaption}
\usepackage{url}
\usepackage{color}

\graphicspath{{images/}}
\DeclareMathOperator{\sigm}{\sigma}
\DeclareMathOperator*{\argmax}{argmax}
\newcommand{\wi}{s}
\newcommand{\wil}{S}
\newcommand{\hi}{k}
\newcommand{\vi}{j}

\newcommand{\all}{\bullet}
\newcommand{\mypar}[1]{\noindent \textbf{#1:}}
\usepackage{lineno}
\begin{document}

\mainmatter  

\title{Rotation-Invariant Restricted Boltzmann Machine Using Shared Gradient Filters}

\titlerunning{Explicit Rotation-Invariant RBM}

%
%
\author{Mario Valerio Giuffrida$^{1,2}$, \and Sotirios A. Tsaftaris$^{1,2}$}

%
\authorrunning{M.V. Giuffrida, S. A. Tsaftaris}

\institute{
	$^1$ IMT Scuola Alti Studi Lucca, PRIAn, Italy.\\
	$^2$ School of Engineering, University of Edinburgh, UK.
	\\
	\url{valerio.giuffrida@imtlucca.it}\\
	\url{s.tsaftaris@ac.ed.uk}}

%
%

\toctitle{Lecture Notes in Computer Science}
\tocauthor{Authors' Instructions}
\maketitle

\begin{abstract}
Finding suitable features has been an essential problem in computer vision. 
We focus on Restricted Boltzmann Machines (RBMs), which, despite their versatility, cannot accommodate transformations that may occur in the scene. 
As a result, several approaches have been proposed that consider a set of transformations, which are used to either augment the training set or transform the actual learned filters. In this paper, we propose the \textit{Explicit Rotation-Invariant Restricted Boltzmann Machine}, which exploits prior information coming from the dominant orientation of images.  
Our model extends the standard RBM, by adding a suitable number of weight matrices, associated with each dominant gradient. We show that our approach is able to learn rotation-invariant features, comparing it with the classic formulation of RBM on the MNIST benchmark dataset. Overall, requiring less hidden units, our method learns compact features, which are robust to rotations. 
\end{abstract}

\section{Introduction}

\label{sec:intro}

It is widely known that a crucial problem in image understanding is to find suitable features for the task at hand. Hand-crafted descriptors were able to provide adequate representations, but they rely on specific structures in the scene and could not accommodate certain nuisance factors properly. Hence, extensive efforts in learning image representations have been done in the past years, demonstrating that machine learning approaches are able to outperform hand-crafted descriptors \cite{Wei2015}. Examples of learned features are e.g. vocabulary learning \cite{Csurka2004}, sparse coding \cite{Lee2006}, Gaussian mixture models \cite{Agarwal2006}, neural networks \cite{Arel2010}. 

Neural networks (NNs) are graphical models, where nodes in a graph are connected with weighted connections and parameters are determined via optimisation algorithms.
The \textit{Restricted Boltzmann Machine} (RBM) has recently gained popularity, mainly because of its applications to deep learning \cite{Arel2010,Hinton2006}. RBM is a generative NN constituted by a bipartite graph, which sides are referred to \textit{visible layer} and \textit{hidden layer} respectively. The set of parameters within the RBM are optimised via the \textit{Contrastive Divergence} (CD) algorithm \cite{Hinton2002}. Although RBMs can achieve satisfactory results \cite{Coates2011}, their use in shallow networks (namely few layers) cannot accommodate complex variability occurring in the scene \cite{Shou2013}. To this end, the \textit{Deep Belief Network} (DBN) was proposed in \cite{Larochelle2007}, which is constituted by several stacked RBMs. Albeit DBN have been shown to achieve some translation invariance, they may not well accommodate other nuisance factors (e.g. rotation).

\begin{figure}[t]
	\centering
	\includegraphics[width=0.9\textwidth]{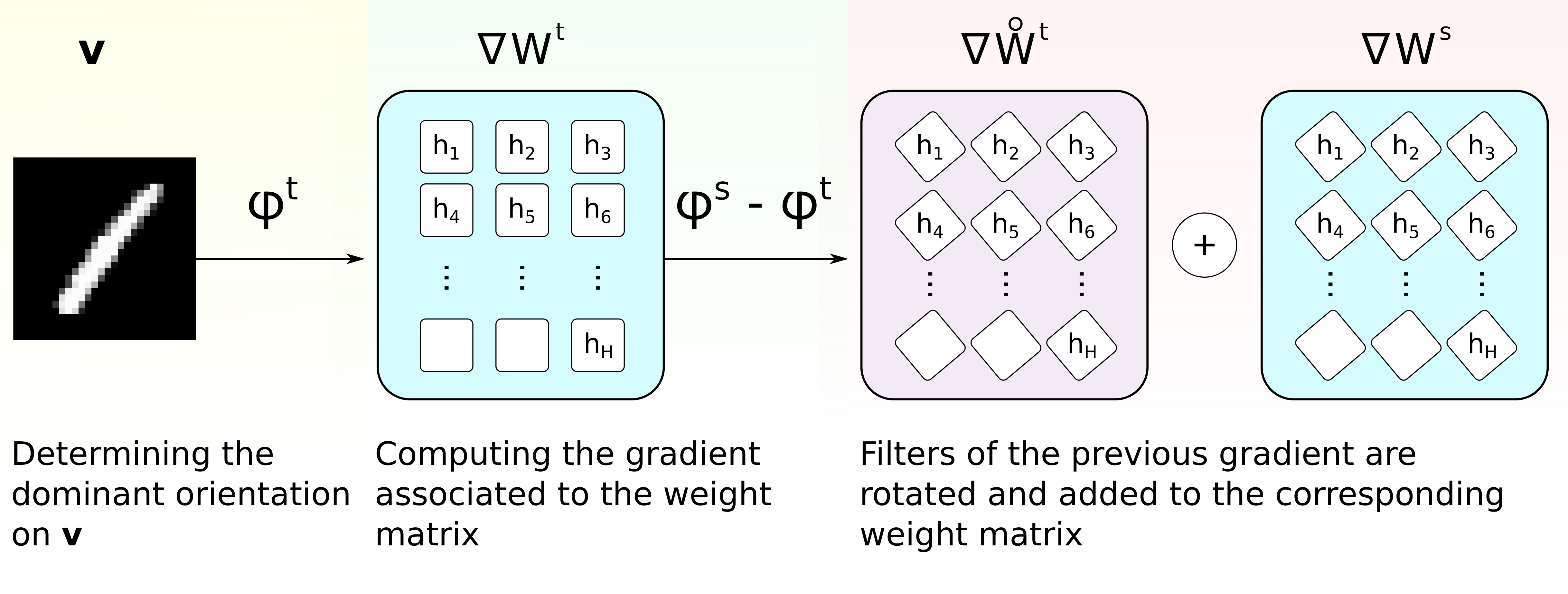}
	\caption{The dominant orientation $\varphi^t$ is determined for the provided image and is used to compute the gradient $\nabla W^{(t)}$. The contribution of this gradient is shared amongst the other weight matrices  $\nabla W^{(\wi)}$, $\wi=1,2,\ldots,\wil$, $t\neq \wi$, rotating the learned filters by the angle $\varphi^\wi - \varphi^t$ to generate the $\nabla \mathring{W}^{(t)}$ term.}
	\label{fig:erirbm}
\end{figure}

In fact, several modifications of the original RBM formulation have been recently proposed, achieving certain transformation invariance. In \cite{Sohn2012}, a transformation invariant RBM is proposed, where images are subjected to a predefined set of transformations. In \cite{Kivinen2011} an RBM that learns equivariant features is proposed, whereby adding a new variable to be inferred within the hidden units, this variable is then used to rotate learned weights accordingly. In \cite{Schmidt2012}, a rotation (invariant) Convolutional RBM is proposed. The marginal probability of RBM is extended with a Markov Random Field, including transformed versions of input images. In \cite{Shou2013}, an additional step of the backpropagation algorithm used to train DBN is introduced, where the weights are transformed and the entire network is trained again. In \cite{Cheng2013}, the authors propose an RBM where input images are divided into non-overlapping blocks. Then, patches are extracted on SIFT keypoints \cite{Lowe1999} and subsequently rotated and scaled accordingly. Despite their progress, the aforementioned methods share the following drawbacks: either they are limited to the set of transformations considered within the model, or they involve deep networks in the hope of learning better transformation invariant features \cite{Kivinen2011,Shou2013,Sohn2012}, albeit increasing computational demand. 

In this paper instead we present the \textit{Explicit Rotation-Invariant Restricted Boltzmann Machine} (ERI-RBM), which can model the nuisance caused by rotated versions of the same pattern, without actually applying any transformation to the data. Our method considers a set of weight matrices (similar concept as in C-RBM \cite{Lee2008}) and each sample is provided to the visible layer with its dominant orientation \cite{Cheng2013}. This information is used to select a particular weight matrix during the Gibbs sampling to compute gradients of parameters. The contribution given by the new update gradients is shared among the other weight matrices, rotating the filters accordingly \cite{Shou2013} (cf. \Cref{fig:erirbm}). Experiments on MNIST-rot show superior performance to several baseline benchmarks and a recent method from the literature. 

Our contributions are multi-fold: (i) rotation is treated explicitly, without rotating the image patterns, in contrast to for example \cite{Sohn2012}; (ii) we adopt a shallow model using a limited amount of additional weight matrices, instead of deep architectures \cite{Lee2009}; (iii) we share the contribution coming from a weight matrix with the other ones, rotating the learned filters by suitable angles.

This paper is organised as follows. \Cref{sec:rirbm} describes the proposed Explicit Rotation-Invariant Restricted Boltzmann Machine. In \Cref{sec:res}, we present experimental results, whereas \Cref{sec:concl} concludes the manuscript.

\section{Explicit Rotation-Invariant RBM (ERI-RBM)}
\label{sec:rirbm}

In this section, we discuss how to embed the concept of rotation-invariance explicitly in the RBM formulation. Since input patterns are images, we will assume that neurones in the visible layer are arranged in matrix form of size $w \times h = d$, width and height respectively. Each row in the weight matrix $W$, connecting visible units to hidden units, is a $d$-dimensional vector. Therefore, each row in $W$ can also be arranged in matrix form of size $w \times h$. Henceforth, we will refer to rows in the weight matrix $W$ as \textit{learned filters} and rows in $\nabla W$ as \textit{update filters}, which is the gradient computed during the Contrastive Divergence algorithm.

\subsection{Proposed model}
Let $\Phi$ be a set of evenly distanced angles $\Phi = \left\lbrace \varphi_1, \varphi_2, \ldots, \varphi_\wil \right\rbrace$, such that for any $i\leq j \Longrightarrow \varphi_i \leq \varphi_j$. In our model, we augment the number of weight matrices $W \in \mathbb{R}^{H\times V \times \wil}$, such that every angle $\varphi_s$ is associated to a matrix $W^{(s)}$. Here, $H$ is the number of hidden units, $V$ the number of visible units, and $S$ is the number of angles. In addition, each weight matrix has  an associated bias vector $\boldsymbol{b}^{(\wi)}$. Hence, we rewrite the energy function characterising the standard Restricted Boltzmann Machine formulation as follows:

\begin{equation}
\label{eq:energy_edit}
E(\boldsymbol{v},\boldsymbol{h}; \wi) = -\boldsymbol{h}^T W^{(\wi)} \boldsymbol{v} - \boldsymbol{c}^T \boldsymbol{v} - \left[\boldsymbol{b}^{(\wi)}\right]^T \boldsymbol{h},
\end{equation}

\noindent where $W^{(\wi)}$ is the $\wi$-th weight matrix, $\boldsymbol{b}^{(\wi)}$ is the bias vector for the hidden layer associated to $W^{(\wi)}$, with $\wi=1,2,\ldots,\wil$, and $\boldsymbol{c}$ is the bias vector for the visible layer. The index $s$ is uniquely determined on each input image $\boldsymbol{v}$, and will be discussed thoroughly in \Cref{subsec:angle}. Because of the modification in \eqref{eq:energy_edit}, all the equations involved in the CD algorithm have to be rewritten. Specifically, the conditional probabilities become:

\begin{equation}
\label{eq:hgivenx_edit}
p({h_\hi = 1}|\boldsymbol{v};\wi) = \sigm{\left( b^{(\wi)}_\hi + \boldsymbol{W^{(\wi)}_{\hi,\all}} \boldsymbol{v} \right)},
\end{equation}

\begin{equation}
\label{eq:xgivenh_edit}
p({v_\vi = 1}|\boldsymbol{h};\wi) = \sigm{\left( c_\vi + \boldsymbol{h}^T \boldsymbol{W^{(\wi)}_{\all, \vi}} \right)}.
\end{equation}

During the optimisation algorithm, an image $\boldsymbol{v}$ with dominant orientation $\varphi_s$ is provided to the Gibbs sampling. After a sufficient number of alternating computations of \eqref{eq:hgivenx_edit} and \eqref{eq:xgivenh_edit}, the gradient $\nabla W^{(\wi)}$ can be computed, whose contribution is shared with the remaining matrices in $W$. To update $\nabla W^{(t)}$, $1\leq t \leq \wil$, $t \neq \wi$, we transform the update filters in $\nabla W^{(\wi)}$ which are then added to the $t$-th gradient. Specifically, since we can represent rows in $\nabla W^{(\wi)}$ as images, they can be rotated by an angle $\theta = \phi_t - \phi_\wi$. Therefore, we define a new \textit{shared update filter} term $\nabla \mathring{W}^{(t)}$, such that

\begin{equation}
\label{eq:nablaringW}
\nabla \mathring{W}^{(t)} = R_{\theta} (\nabla W^{(\wi)}) \equiv  \left(\begin{matrix}
R_{\theta}\left(\boldsymbol{\nabla W^{(\wi)}_{1,\bullet}}\right)\\
R_{\theta}\left(\boldsymbol{\nabla W^{(\wi)}_{2,\bullet}}\right)\\
\vdots \\
R_{\theta}\left(\boldsymbol{\nabla W^{(\wi)}_{H,\bullet}}\right)\\
\end{matrix} \right).
\end{equation}

\noindent 
where $R_\theta = [ \cos\theta \; -\sin\theta ; \sin\theta \; \cos\theta]$ defines the 2D rotation matrix by an angle $\theta$. This operation may generate filters bigger than the input layers and we crop them such that the filter size remains $w\times h$. At this point, the final expression for the gradient $\nabla W^{(\wi)}$ is updated as follows:

\begin{equation}
\label{eq:new_nabla}
\nabla W^{(\wi)} :=  \nabla W^{(\wi)} + \nabla \mathring{W}^{(\wi)}.
\end{equation}

 Note that \eqref{eq:new_nabla} will be utilised within the Stochastic Gradient Descent step of the CD algorithm. Therefore, $\nabla W^{(\wi)}$ will be multiplied by a learning rate $\eta$ that typically has values set in the order of $10^{-3}$ (further details are discussed in \cite{Hinton2012}). Hence, any side effects originating from pixel interpolation are minimised, precisely because of the small $\eta$. Gradients $\boldsymbol{\nabla b}^{(\wi)}$ are computed as described in \cite{Hinton2002}, using samples $\boldsymbol{v}$ with the associated dominant orientation $\varphi_\wi$.

\subsection{Finding the dominant angle and corresponding $\wi$ index}
\label{subsec:angle}

\begin{figure}[t]
	\centering
	\includegraphics[width=0.9\textwidth]{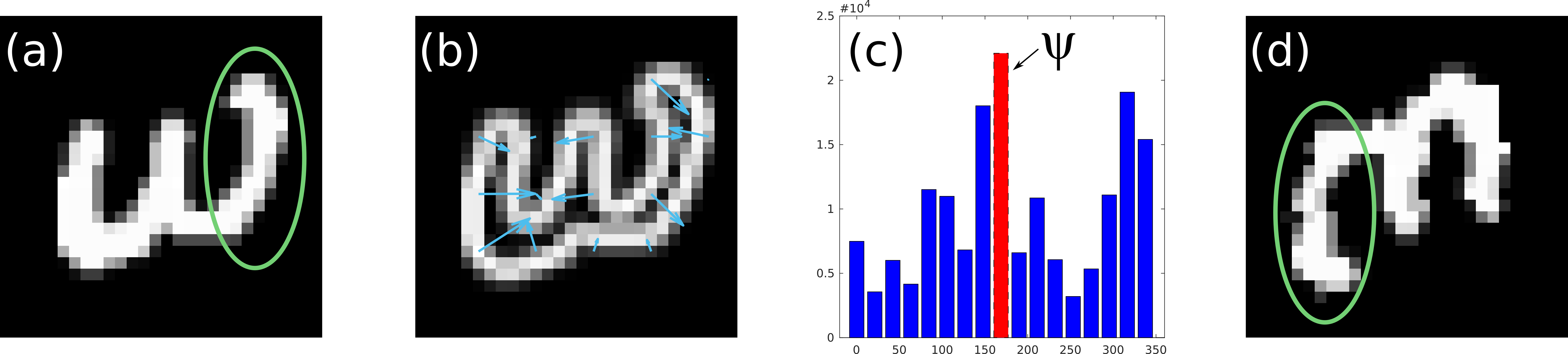}
	\caption{Computation of the dominant orientation for a sample image taken from the MNIST dataset: (a) original sample, (b) gradients of the image, (c) histogram of oriented gradients with highlighted mode $\psi$, (d) sample rotated by $\psi$ degree. The region marked by a green ellipse corresponds to the same portion of the number 3 in the original and rotated image. Observe the differences due to image interpolation introduced during rotation.}
	\label{fig:dom_rot}
	\vspace{-1em}
\end{figure}

Each image $\boldsymbol{v}$ is associated to an angle $\varphi_\wi$, determined by the histogram of oriented gradients from $\boldsymbol{v}$ \cite{Dalal2005}.  Derivatives along the $x$ and $y$ directions are computed and the angle of each gradient vector can be determined. All the vectors are accumulated into a histogram with $S$ bins and the angle $\psi$ with the highest frequency is found. Formally, the index $\wi = \argmax_j \varphi_j$, such that $\varphi_j \leq \psi$, $\varphi_j \in \Phi$. \Cref{fig:dom_rot} shows graphically those steps: from the original image pattern (a), derivatives are computed using Sobel filters (b). Subsequently, we build the weighted histogram of oriented gradients and the angle with the highest frequency $\psi$ is selected (c). We highlight in red the $9$-th bin of the histogram, hence $\wi = 9$ for the illustrated example. In (d) we report a rotated version of the sample image by $\psi$ degree to show the deleterious effect of image interpolation. 

Since strong edges near image boundaries may bias the estimation of the dominant gradient, the magnitude of the corresponding vectors is weighted with a Gaussian kernel, with  $\sigma = \frac{min\{w,h\}}{5}$ (width and height of $\boldsymbol{v}$ respectively), such that central gradients contribute more than those at the boundaries. (We found this value covers evenly the entire image without exceeding its size.)

\section{Experimental results}
\label{sec:res}

\mypar{Setup} We used the MNIST-rot dataset\footnote{Available at \url{http://www.iro.umontreal.ca/~lisa/twiki/bin/view.cgi/Public/DeepVsShallowComparisonICML2007}} \cite{Larochelle2007}, containing $10,000$ images for training, $2,000$ for validation, and $50,000$ for testing. This dataset is derived from the MNIST dataset, where samples were rotated by random angles. To enable comparison with other methods, for consistency, we kept this dataset splitting, and we did not perform cross-validation (that could have provided variances for statistical analysis). Since each image contains several non-zero entries close to 0, we threshold them at a value $\tau=0.3$. We compare ERI-RBM with several informative baselines and a recent invariant method. \textit{Classical RBM:} We trained a standard Bernoulli Restricted Boltzmann Machine and compared results with our Explicit Rotation-Invariant RBM. \textit{Dominant RBM (D-RBM):} We built a simplified model that learns an RBM for each dominant orientation, splitting the training set into $S$ partitions, associated to a different RBM (ie., we have $S$ independent RBMs). \textit{Oriented RBM (O-RBM):} We pre-process the dataset by aligning all images according to their dominant orientation to a reference orientation and train a single RBM. \textit{TI-RBM}: We also compared with the method in \cite{Sohn2012}, using the authors implementation\footnote{Available at \url{https://github.com/kihyuks/icml2012_tirbm}}. Extracted features are provided to the following classifiers: linear and RBF SVM \cite{Vapnik1998}, softmax \cite{Hastie2009}, and K-NN \cite{Dasarathy1991}. 

\begin{table}[t]
	\centering
	\caption{Testing accuracies of standard RBM, Dominant RBM, Oriented RBM, TI-RBM  \cite{Sohn2012}, and our proposed ERI-RBM.} 
	\label{tab:res1}
	
	\begin{tabularx}{\textwidth}{@{}lcccc@{}}
		\toprule
		\multicolumn{1}{l}{}           & \textbf{\begin{tabular}[c]{@{}c@{}}RBF SVM\\ $C=10, \gamma=0.1$ \end{tabular}} & \textbf{\begin{tabular}[c]{@{}c@{}}Linear SVM\\ $C=0.1$\end{tabular}} & \textbf{Softmax} & \textbf{\begin{tabular}[c]{@{}c@{}}K-NN\\ K=3\end{tabular}} \\ \midrule
		\textbf{RBM (H=100)}           & 87.37\%                                                                & 59.27\%             & 57.80\%          & 82.69\%                                                     \\ \midrule
		\textbf{D-RBM (H=100, S=4)}    & 83.44\%                                                                & 58.95\%             & 56.80\%          & 78.84\%                                                     \\
		\textbf{D-RBM (H=100, S=9)}    & 79.18\%                                                                & 53.62\%             & 50.76\%          & 73.56\%                                                     \\
		\textbf{D-RBM (H=100, S=18)}   & 69.84\%                                                                & 49.20\%             & 46.58\%          & 63.61\%                                                     \\ \midrule
		\textbf{O-RBM (H=100 S=18)} & 87.37\% & 58.99\% & 57.80\% & 82.69\% \\ \midrule

		\textbf{ERI-RBM (H=100, S=4)}  & 78.49\%                                                                & 60.27\%             & 58.31\%          & 74.97\%                                                     \\
		\textbf{ERI-RBM (H=100, S=9)}  & 91.27\%                                                                & 74.87\%             & 73.02\%          & 88.48\%                                                     \\
		\textbf{ERI-RBM (H=100, S=18)} & \textbf{92.08}\%                                                                & \textbf{77.69}\%             & \textbf{75.84}\%          & \textbf{89.34}\% \\ \midrule   
		\textbf{TI-RBM \cite{Sohn2012} (H=100, S=18)} & 80.63\%                                                      & 69.10\%             & 68.20\%          & 73.60\%                                                     \\ \bottomrule
	\end{tabularx}
	\vspace{-1em}
\end{table}

\mypar{Parameters} We set the number of hidden units to $H=100$, while progressively increased the number of bins $S$, used to generate the histogram of orientations. Following the instructions in \cite{Hinton2012}, we set the learning rate $\eta = 10^{-3}$, the Contrastive Divergence algorithm is iterated up to 200 epochs, and a constant momentum $\alpha = 0.9$ was used. The parameters for SVM were found using logarithmic grid search and best values are reported in \Cref{tab:res1}. We set arbitrary $K=3$ for the K-NN, using the Euclidean distance as metric. For TI-RBM \cite{Sohn2012}, a set of $K=S$ transformations are considered, which is each associated with an array of $H$ hidden units, while a single weight matrix $W$ is considered. The final representation used during inference is obtained by max-pooling. To make the comparison to ERI-RBM fair, for TI-RBM the sparsity term was disabled, and we set the number of hidden units to $H=100$. 

\begin{figure}[t]
	\centering
	\includegraphics[width=0.9\textwidth,trim={0 1.4cm 0 0},clip]{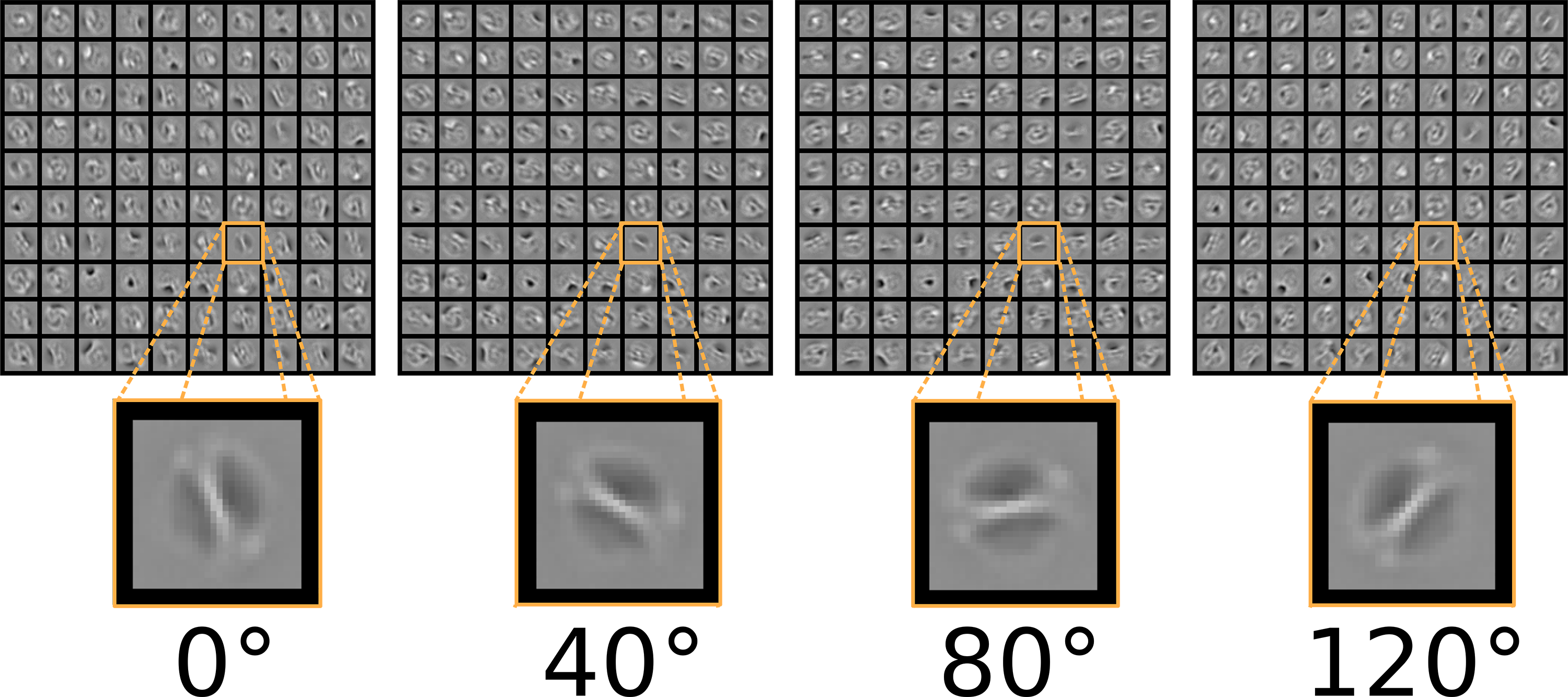}
	\caption{Filters learned by our ERI-RBM at $S=9$. We highlight a filter that appears at rotations $0^\circ$, $40^\circ$, $80^\circ$, and $120^\circ$, showing that our model learns rotation-invariant filters. The remaining weight matrices are omitted for brevity.}
	\label{fig:W}
	\vspace{-1.5em}
\end{figure}

 \mypar{Discussion} We report our results in \Cref{tab:res1} and we noticed that nonlinear SVM gave the best performance in all the cases. The baseline is given by RBM with an accuracy of $87\%$. Tests using D-RBM show a gradual loss of accuracy as the number of dominant orientations $S$ is increased. This behaviour can be attributed to the lack of information sharing amongst the RBMs, since they were each trained independently with less data (per RBM). Overall, our proposed model outperforms the baseline RBM ($S\geq 9$). At $S=4$, ERI-RBM has a loss of performance, because of the coarse quantization of the $2\pi$ space: angles $0^\circ$, $90^\circ$, $180^\circ$, and $270^\circ$ will have orthogonal rotations when shared update filters are computed for neighbour matrices, causing the propagation of sharp rotations that do not contribute much. As the number of $S$ increases, ERI-RBM has a $+13\%$ of improvement, showing that our model is able to learn rotation-invariant features. This is also displayed in \Cref{fig:W}, showing learned filters when $S=9$. O-RBM shows no improvement compared to RBM, demonstrating that the contribution provided by the shared update filters increases the discriminative power of the final representation. Note that we also trained classical RBM with $H=1000$, noticing an improvement of $2\%$, still lower than ERI-RBM. Finally, using the same experimental setup, ERI-RBM outperformed \cite{Sohn2012} by $+12\%$ in testing accuracy. (These results are different from those reported in \cite{Sohn2012} since sparsity is not present and we used less units.). Our approach does rely on the determination of orientation, which could be seen as a limitation. Preliminary results (not shown for brevity), obtained by artificially perturbing the orientation estimate, show that we are tolerant to such errors up to $\pm 4$ bins off on the original estimate. This remains to be confirmed in images with cluttered background.

\section{Conclusions}
\label{sec:concl}

In this paper we proposed the \textit{Explicit Rotation-Invariant Restricted Boltzmann Machine} (ERI-RBM). Current approaches do not address the problem of rotation-invariance directly, but use a predefined set of transformations to transform either the input images \cite{Schmidt2012,Sohn2012} or the learned filters \cite{Kivinen2011,Shou2013}. We were inspired by these approaches to modify the RBM learning process, such that to learn invariant features without taking into account all possible transformations, which is demanding and may propagate noise due to pixel interpolations.

Our ERI-RBM utilises the dominant gradient of input images in order to select the best set of filters to optimise. We find the corresponding gradients efficiently and update the filters in a process where information is shared across the different filters, minimising thus any effects of interpolation.  Overall, our model learns rotation-invariant features and achieves an accuracy of $92\%$ in the MNIST-rot dataset. Comparisons with several baselines and approaches from the literature showed superior performance in a common experimental setup. Moreover, comparing to the deep architecture of \cite{Gens14} and the results on MNIST-rot, ERI-RBM reached similar performance using just 100 of hidden units compared to the 500 in \cite{Gens14}. In conclusion,  ERI-RBM is able to learn rotation-invariant features in an unsupervised fashion, with a reduced number of hidden units, within a shallow network.

\section*{Acknowledgements}
We thank NVIDIA corporation for providing us a Titan X GPU.

\bibliographystyle{splncs03}
\bibliography{ref}

\begin{thebibliography}{10}
\providecommand{\url}[1]{\texttt{#1}}
\providecommand{\urlprefix}{URL }

\bibitem{Agarwal2006}
Agarwal, A., Triggs, B.: Computer Vision -- ECCV: 9th European Conference on
  Computer Vision, chap. Hyperfeatures -- Multilevel Local Coding for Visual
  Recognition, pp. 30--43. Springer Berlin Heidelberg (2006)

\bibitem{Arel2010}
Arel, I., Rose, D.C., Karnowski, T.P.: {Deep Machine Learning - A New Frontier
  in Artificial Intelligence Research}. IEEE Computational Intelligence
  Magazine  5(4),  13--18 (2010)

\bibitem{Cheng2013}
Cheng, D., Sun, T., Jiang, X., Wang, S.: {Unsupervised feature learning using
  Markov deep belief network}. In: 2013 IEEE International Conference on Image
  Processing. pp. 260--264. No. 20120073110053, IEEE (2013)

\bibitem{Coates2011}
Coates, A., Arbor, A., Ng, A.Y.: {An Analysis of Single-Layer Networks in
  Unsupervised Feature Learning}. {AISTATS } pp. 215--223 (2011)

\bibitem{Csurka2004}
Csurka, G., Dance, C.R., Fan, L., Willamowski, J., Bray, C.: {Visual
  categorization with bags of keypoints}. Proceedings of the ECCV International
  Workshop on Statistical Learning in Computer Vision pp. 59--74 (2004)

\bibitem{Dalal2005}
Dalal, N., Triggs, B.: {Histograms of oriented gradients for human detection}.
  Proceedings of the IEEE CVPR  1,  886--893 (2005)

\bibitem{Dasarathy1991}
Dasarathy, B.: Nearest neighbor ({NN}) norms: nn pattern classification
  techniques. IEEE Computer Society Press (1991)

\bibitem{Gens14}
Gens, R., Domingos, P.M.: {Deep Symmetry Networks}. In: {NIPS}. pp. 2537--2545.
  Curran Associates, Inc. (2014)

\bibitem{Hastie2009}
Hastie, T., Tibshirani, R., Friedman, J.: {The Elements of Statistical
  Learning}, Springer Series in Statistics, vol.~1. Springer New York, 2nd edn.
  (2009)

\bibitem{Hinton2012}
Hinton, G.: {A Practical Guide to Training Restricted Boltzmann Machines}.
  Springer Berlin Heidelberg, 2nd edn. (2012)

\bibitem{Hinton2002}
Hinton, G.E.: {Training products of experts by minimizing contrastive
  divergence.} Neural computation  14(8),  1771--1800 (2002)

\bibitem{Hinton2006}
Hinton, G.E., Osindero, S., Teh, Y.W.: {A fast learning algorithm for deep
  belief nets.} Neural computation  18(7),  1527--54 (2006)

\bibitem{Kivinen2011}
Kivinen, J.J., Williams, C.K.I.: {Transformation Equivariant Boltzmann
  Machines}. In: ICANN, vol. 6791, pp. 1--9. Springer (2011)

\bibitem{Larochelle2007}
Larochelle, H., Erhan, D., Courville, A., Bergstra, J., Bengio, Y.: {An
  empirical evaluation of deep architectures on problems with many factors of
  variation}. Proceedings of the 24th ICML pp. 473--480 (2007)

\bibitem{Lee2006}
Lee, H., Battle, A., Raina, R., Ng, A.Y.: {Efficient Sparse coding algorithms}.
  Advances in neural information processing systems pp. 801--808 (2006)

\bibitem{Lee2008}
Lee, H., Ekanadham, C., Ng, A.Y.: {Sparse deep belief net model for visual area
  V2}. Advances in Neural Information Processing Systems pp. 873--880 (2008)

\bibitem{Lee2009}
Lee, H., Grosse, R., Ranganath, R., Ng, A.Y.: {Convolutional deep belief
  networks for scalable unsupervised learning of hierarchical representations}.
  ICML  (2009)

\bibitem{Lowe1999}
Lowe, D.G.: Object recognition from local scale-invariant features. In: ICCV
  (1999)

\bibitem{Schmidt2012}
Schmidt, U., Roth, S.: {Learning rotation-aware features: From invariant priors
  to equivariant descriptors}. Proceedings of the IEEE CVPR pp. 2050--2057
  (2012)

\bibitem{Shou2013}
Shou, Z., Zhang, Y., Cai, H.J.: {A study of transformation-invariances of deep
  belief networks}. In: IJCNN. pp. 1--8. IEEE (2013)

\bibitem{Sohn2012}
Sohn, K., Lee, H.: {Learning Invariant Representations with Local
  Transformations}. Proceedings of the 29th ICML pp. 1311--1318 (2012)

\bibitem{Vapnik1998}
Vapnik, V.: Statistical Learning Theory. John Wiley and Sons (1998)

\bibitem{Wei2015}
Wei, X., Phung, S.L., Bouzerdoum, A.: {Visual descriptors for scene
  categorization: experimental evaluation}. Artificial Intelligence Review
  45(3),  1--36 (2015)

\end{thebibliography}

\end{document}